# A Recent Survey on the Applications of Genetic Programming in Image Processing

Asifullah Khan[*1,2], Aqsa Saeed Qureshi[1], Noorul Wahab[1], Mutawara Hussain[1], Muhammad Yousaf Hamza[3]

[1] Patter Recognition Lab, Department of Computer and Information Sciences, Pakistan Institute of Engineering and Applied Sciences, Nilore 45650, Islamabad

[2] Deep Learning Lab, Centre for Mathematical Sciences, Pakistan Institute of Engineering and Applied Sciences, Nilore 45650, Islamabad

[3] Department of Physiscs and Applied Mathematics, Pakistan Institute of Engineering and Applied Sciences, Nilore-45650, Islamabad

asif@pieas.edu.pk

**Abstract**

Genetic Programming (GP) has been primarily used to tackle optimization, classification, and feature selection related tasks. The widespread use of GP is due to its flexible and comprehensible tree-type structure. Similarly, research is also gaining momentum in the field of Image Processing, because of its promising results over vast areas of applications ranging from medical Image Processing to multispectral imaging. Image Processing is mainly involved in applications such as computer vision, pattern recognition, image compression, storage, and medical diagnostics. This universal nature of images and their associated algorithm, i.e., complexities, gave an impetus to the exploration of GP. GP has thus been used in different ways for Image Processing since its inception.

Many interesting GP techniques have been developed and employed in the field of Image Processing, and consequently, we aim to provide the research community an extensive view of these techniques. This survey thus presents the diverse applications of GP in Image Processing and provides useful resources for further research. Also, the comparison of different parameters used in different applications of Image Processing is summarized in tabular form. Moreover, analysis of the different parameters used in Image Processing related tasks is carried-out to save the time needed in the future for evaluating the parameters of GP. As more advancement is made in GP methodologies, its success in solving complex tasks, not only in Image Processing but also in other fields, may increase. Additionally, guidelines are provided for applying GP in Image Processing related tasks, the pros and cons of GP techniques are discussed, and some future directions are also set.

## 1. Introduction

The sense of vision plays an essential role in the process of human perception. Human vision is restricted only to the visual band of the electromagnetic spectrum, but machine vision covers nearly the whole electromagnetic spectrum, ranging from gamma rays to radio waves[1]. Image processing tries to emulate the capabilities of the human eye and brain in extracting features or segmenting regions. Therefore, image processing is a challenging task in the sense that these algorithms have to be accurate, fast, reliable, as well as robust. Development in image processing has considerably increased with the decline in the prices of computers. Due to its diverse applications, image processing cannot be wholly distinguished from its closely related fields; computer vision and image analysis. This overlapping is because image processing is also involved in both computer vision and image analysis at different levels. In the somewhat restricted definition of image processing, it is a process whose inputs and outputs are images and can be extended to encompass processes that involve techniques of feature extraction from images to identify the individual objects [2].

Different intelligent techniques such as an Artificial Immune System (AIS), Genetic Algorithm (GA), Artificial Neural Network (ANN), Ant Colony Optimization (ACO), and Genetic Programming (GP) have been exploited in the field of image processing. The term "Computational Intelligence" (CI) is a broad term encompassing the different intelligent techniques mentioned above. CI can find optimum/near-optimum solutions to computationally hard problems in a variety of domains [3]. Thus to split the various image processing techniques, we have two broad categories; conventional image processing and CI-based image processing techniques.

This survey focuses on the applications of GP in image processing. GP is one of the promising CI technique that comes under the sub-category of Evolutionary Computation (EC) techniques based on the Darwinian theory of evolution. GP evolves output in the form of a tree or a computer program. Different programs are generated depending on the terminal and function sets used. Other existing CI paradigms do not produce solutions in the form of computer programs, but instead involve specialized structures like weight vectors for neural networks, coefficients for polynomials, chromosome strings in the conventional GA, etc. [4]. GP comes under the umbrella of EC along with GA, Evolutionary Programming, Differential Evolution, and Evolutionary Strategies [5]. GP is a particular form of the GA, which uses a fixed (though variants now exist) length string of bits or real numbers to represent individuals called chromosomes.

In contrast to GA, GP represents individuals as trees that can be evaluated to obtain results. Initially, a population of individuals is randomly generated using a terminal set (which contains constants, argument-less functions, variables) and a function set (e.g., +, -, /, if-else, etc.). Based on their fitness, the individuals are given chances for reproduction and allowed to change via crossover and mutation. Crossover is used to search for an optimal solution, whereas mutation introduces rapid changes in the population and thus helps to avoid trapping in local optima.

GP has gained popularity in applications such as data modeling, symbolic regression, Image and Signal Processing, Medicine, Bioinformatics, Financial Trading, and Industrial Process Control. This popularity of GP is mainly through its flexible nature, generality, almost no requirement of preprocessing, its ability to provide the mathematical expression of the solution, and parallelization.

This survey addresses GP's applicability in image processing and is organized as follows. The background of GP and image processing is described in Section 2. The importance of the review is presented in Section 3. The similarities of the GP approaches in different categories of IP are given in Section 4 and further reviewed in Section 5. Section 6 is about the advantages and disadvantages of using GP in image processing. Section 7 presents guidelines for applying GP in image processing. The comparison and discussions are provided in Section 8, while Section 9 concludes the article.

## 2. Related Concepts

This section briefly describes the very basics of the two subjects of this review; image processing and GP. We also discuss the scope of GP in image processing.

**a) image processing**

Image is a visual representation of an object produced on a surface. Before the invention of paper, images were produced on stones and other materials. In the case of computers, a visual representation of an image is displayed on a monitor, a Liquid Crystal Display, or a multimedia projector. However, for computer storage, these images are defined as two-dimensional matrices of pixel (picture-element) values. These pixel values are the intensity or gray level of the image and can be represented in the form of a function $F(x,y)$, where $x$ and $y$ are spatial coordinates. If the intensity values within an image are finite discrete quantities, then, such an image is a digital image. A pixel of size one byte (8 bits) can represent 256 intensity values from 0 (black) to 255 (white). The values in between this range give different shades, as shown in Figure 1. When values of such a representation are processed/modified in some way, we call it image processing. For example, enhancing the image quality, removing noise, segmenting specific parts, making a comparison with other images, etc., all include processing the image in some way. For the image in Figure 1, if we want to change the center pixel to black, then we change its value from 78 to 0.

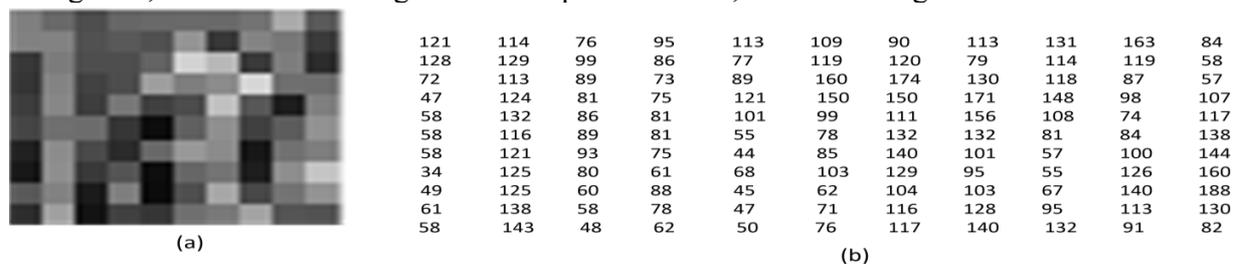

Figure 1: a) a sample image b) its pixel representation

**b) GP**

GP is one of the promising EC techniques and is viewed as a specialization of GA. GP and GA mainly differ in the representation scheme. GA uses strings of bits, integers, or real numbers to represent individuals. In contrast, GP mainly represents individuals as trees and is well suited for mapping functions, model development, nonlinear regression, and other related problems. Koza has pointed out various exciting problems, where GP produced human-competitive results [6]. GP is a domain-independent method and can solve complex problems automatically [7]. Moreover, pioneering works of Koza, Langdon, Poli, and Banzhaf has boosted research in the field of GP [7–12].

Figure 2 depicts the genetic search cycle of EC techniques, where an initial population is generated, and then, the fittest individuals are selected as parents based on some evaluation criterion. In the next step, genetic operators (e.g., crossover, mutation, and reproduction) are applied to produce offsprings. In the last step, fittest individuals are selected as a population for the next generation. The whole search cycle continues after each generation until a termination criterion fulfills, and the best candidate becomes the fittest individual.

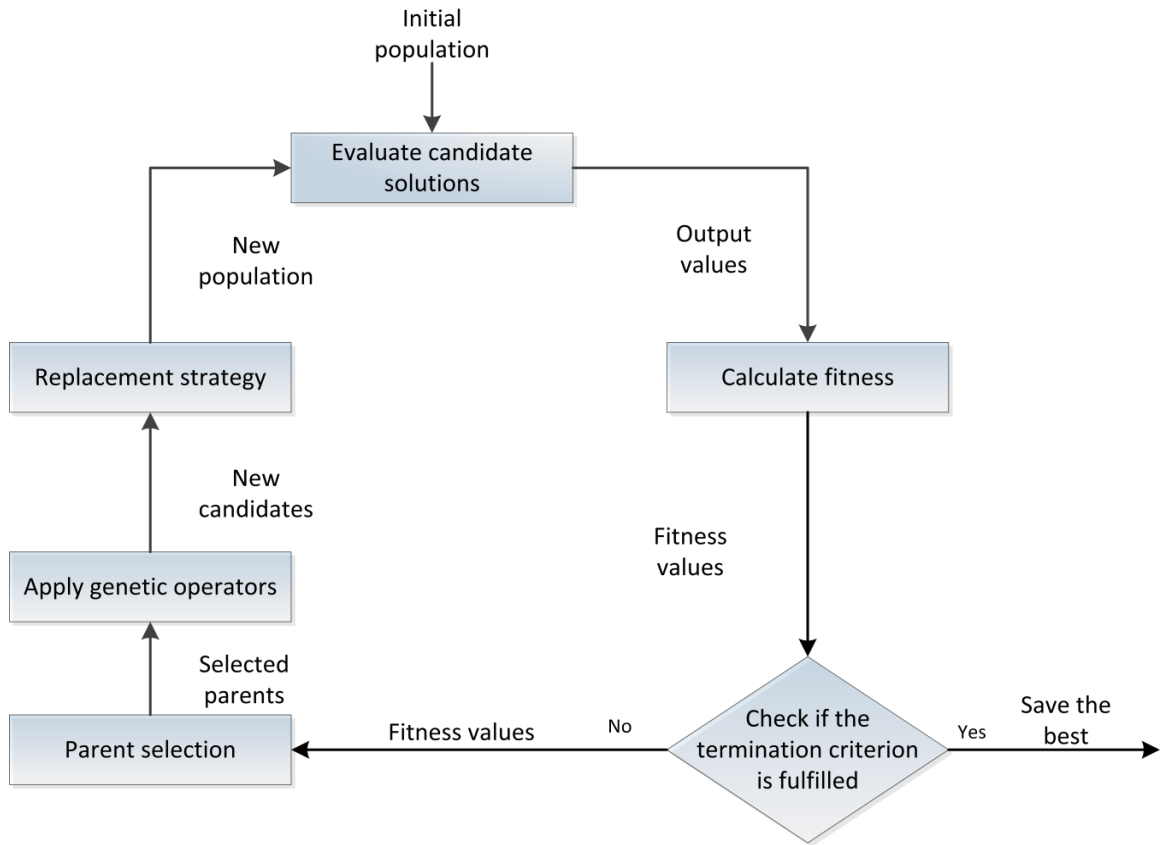

Figure 2: Genetic search cycle of EC techniques

Similarly, Figure 3 depicts the basic flow of GP, in which an initial population is generated randomly. Then parents are selected randomly from this initial population, and different genetic operators are applied. After the application of genetic operators, the selected individuals become part of the next generation. This process is repeated until a termination criterion meets, and finally, the best-evolved GP tree is saved.

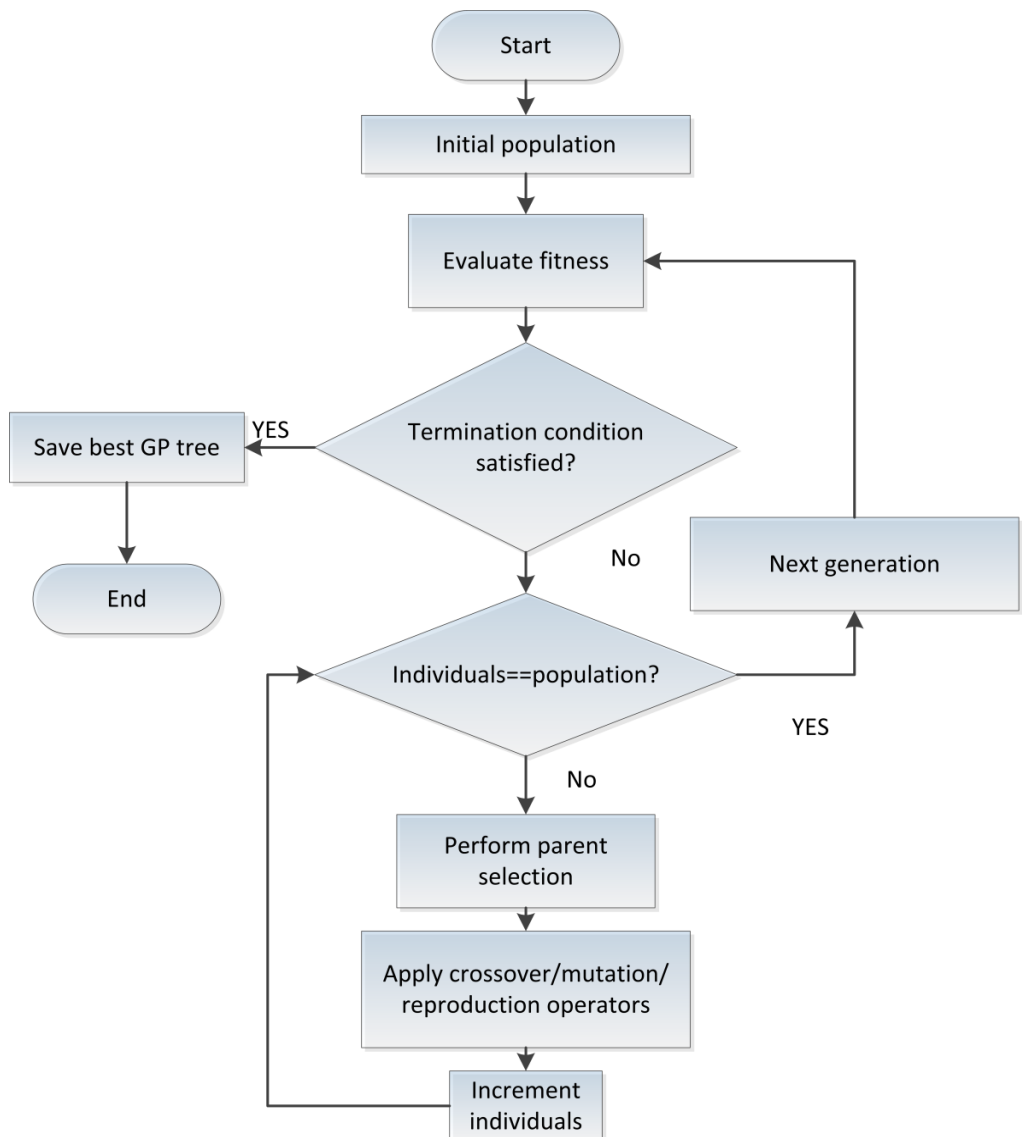
Fig 3: Flowchart of a generic GP

**c) Scope of GP in image processing**
Koza in 1992 introduced GP as a problem-solving paradigm, and since then, it has been applied in many image-related problems. Its expressive power has been utilized in various image processing tasks such as image preprocessing, region analysis, segmentation, object detection, classification, and post-processing. Notably, in the field of medical imaging, it is often applied for classifying cancerous and non-cancerous cells. Moreover, GP has been used for developing accurate classifiers for object detection, classification of medical images, and optical character recognition. Multiobjective GP [13][14][15][16] is also widely used for image processing related problems in which the optimization of more than one objective function is required. Most of the time, the objectives that are needed to optimize are conflicting. For example, in image watermarking techniques, the objective usually is to increase both the imperceptibility and payload of the watermarked image. However, there is always a tradeoff between the two objectives. Similarly, in military-related applications, Similarly, GP is used in military-related applications such as detecting objects, and analysis of Satellite and infrared images. Besides medical and military applications, GP is also employed in other exciting fields such as environmental studies, exploration, crop production, and image indexing [7].

## 3. Importance of the Review

Due to the rapid increase in the availability of images and videos over the last few years, GP has been successfully applied to many image processing applications. In this regard, assessing the prospects of GP in the field of image processing will be a useful guide for researchers. Generally, the performance of algorithms related to segmentation, edge detection, enhancement, and classification related problems suffer if the images are blurred. In this situation, GP can evolve suitable filters so that images are filtered before applying any image processing task. Owing to the importance of GP in image processing, GP based methodologies have been consistently evolving, and new ideas and techniques have been proposed. This survey can help to explore GP related approaches in different areas of image processing. The pros and cons of GP are discussed for practical purposes and further research. Additionally, the techniques presented in this article highlights many aspects of GP.

## 4. Terminologies used in GP

In this section, the various terminologies associated with GP are discussed. The GP techniques applied in different fields of image processing (presented in this article) are different in terms of their domains. Still, they do share some similarities in solving the problems.

**Representation:** In most of the approaches, GP individuals are represented as a tree structure. Moreover, the linear representation for GP, which is also constructed using functions and terminals, is employed in a few works[17].

**Function Set:** The function set is chosen according to the problem at hand. For example, for regression related problems, the function set might comprise of arithmetic operations (*,%,+,-). Similarly, for image processing applications, a specialized function set, according to the nature of the problem domain, may be used.

**Terminal Set:** The terminals like functions, also do not have any specific predefined set. The GP terminal set is comprised of variables (also called program input), constants, or random inputs. In the case of image processing applications, mostly raw pixel values are used as terminals.

**Fitness Function:** Function and terminal sets, which are used to express a GP tree, also define the search space that GP explores during the search process. Fitness function measures how good or bad is a specific region within the search space. Different fitness functions, depending on the nature of the problem, have been used as an evaluation measure during the search process, such as Root-Mean-Square-Error (RMSE), Peak-Signal-to-Noise Ratio (PSNR), Accuracy, Area-Under the Receiver Operating Characteristic (AUC-ROC) curve, etc.

**Initial Population*:* If prior knowledge about the properties of the desired solutions is not known, then the initial individuals are generated randomly. Moreover, other methods initialize the population with the help of a seed [4].

**Selection Method*:* In GP evolution cycles, mainly two types of selection methods are used, i.e., parent selection and survivor selection. There are different types of selection methods, but tournament selection is the widely used selection mechanism. In parent selection, individuals having higher fitness are selected as parents for the next generation. Whereas, survivor selection is performed on individuals who are produced from selected parents.

**Genetic Operators:** Different genetic operators (crossover, mutation, reproduction, etc.) are used for the generation of offspring to introduce diversity among the individuals of the population.

## 5. Category-wise image processing Applications of GP

This section presents the different GP techniques applied in various fields of image processing, such as image enhancement, compression, segmentation, retrieval, classification, and registration.

### 5.1. GP in Image Enhancement

Images can be enhanced to improve their visual appearance. Enhanced images can further improve the image processing related tasks such as image segmentation, object detection, and recognition. However, image enhancement for one application may not be the right candidate for another application, and this means that image enhancement has different semantics for different applications.

Different image enhancement techniques can be carried out either in the original (spatial) or transformed (frequency) domain. In the original domain, the operations are carried out directly on the pixels. In contrast, in the case of the frequency domain, firstly, the image is transformed in the frequency domain, and then the enhancement is carried out. Sometimes the desired objects that need to be detected (called the region of interest) are emphasized during the enhancement step to help perceive them. For example, to make the process of object extraction easy, an image can be enhanced by decreasing the similarities between the specified object and the background.

In an exciting work, Poli et al. [18]used a pseudo color transformation that utilized GP and developed a program for image enhancement. Similarly, Wang et al. [19] used GP algorithms to evolve morphological operations that converted a binary image into the desired image, which contained only the required features. In Wang's approach, the automatic evaluation mechanism enabled the GP algorithm to generate practical morphological algorithms. On the other hand, Khan et al. [20] proposed a GP based hybrid filter that assisted in reducing the region noise. Their method helped to preserve the details related to edges and structure of the region. Block diagram of Khan's technique is shown in Figure 4. It comprises of two phases. In the first phase, features were extracted from the noisy MRI images using three different types of filters. The extracted features were concatenated to form a feature vector. This feature vector was then used in the second phase to train a GP module. After the training phase, the best evolved GP expression was utilized to check the effectiveness of the proposed technique on new images.

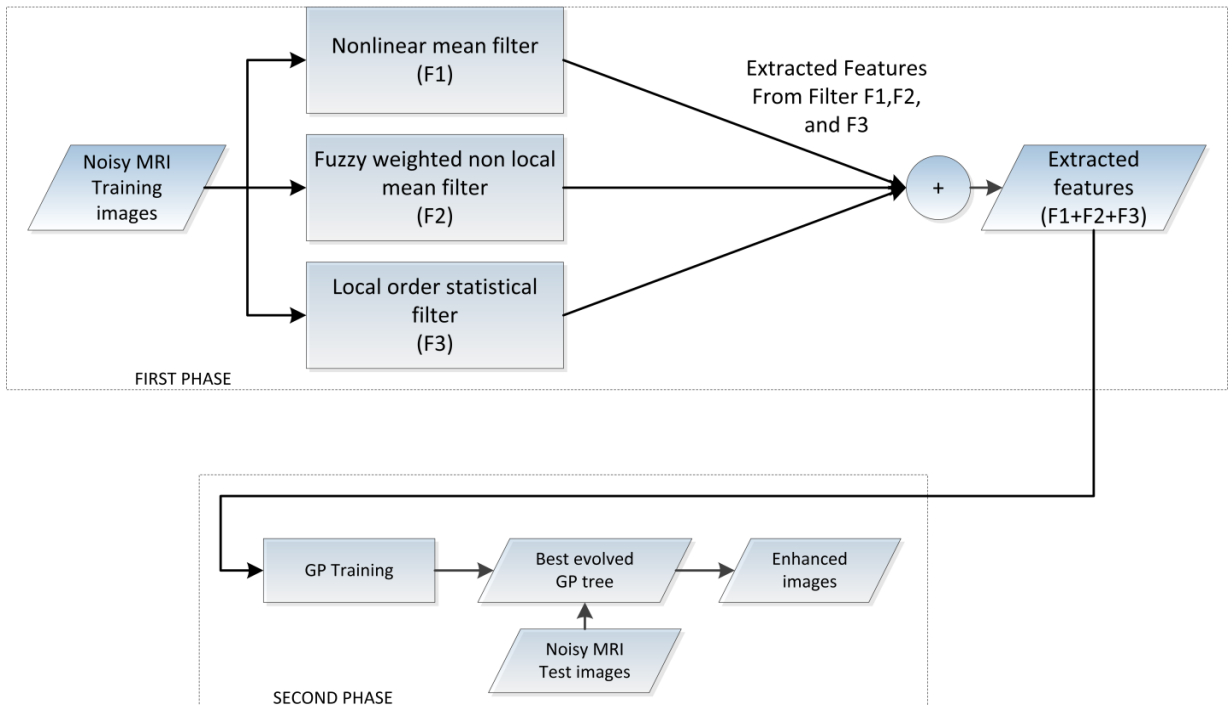

Figure 4: GP based Image enhancement technique by Khan et al. [20]

## 5.2. GP in Image Restoration

During the process of acquiring and transmitting or capturing digital images, the image quality might degrade. Different techniques can be applied to restore the original image. In the case of image restoration, the cause of degradation is either known or unknown. Figure 5 shows the case when the cause of degradation is known. In this case, the original image can be restored using prior knowledge. In a case where there is no such information, then the degraded function can be estimated by image observation, experimentation, or modeling. In literature, blind deconvolution and image denoising based methods are reported for restoring the original image. Restoration by an estimated degradation function is sometimes called blind deconvolution. Whereas in the case of image denoising, spatial or frequency domain filters are used for the restoration of the original image.

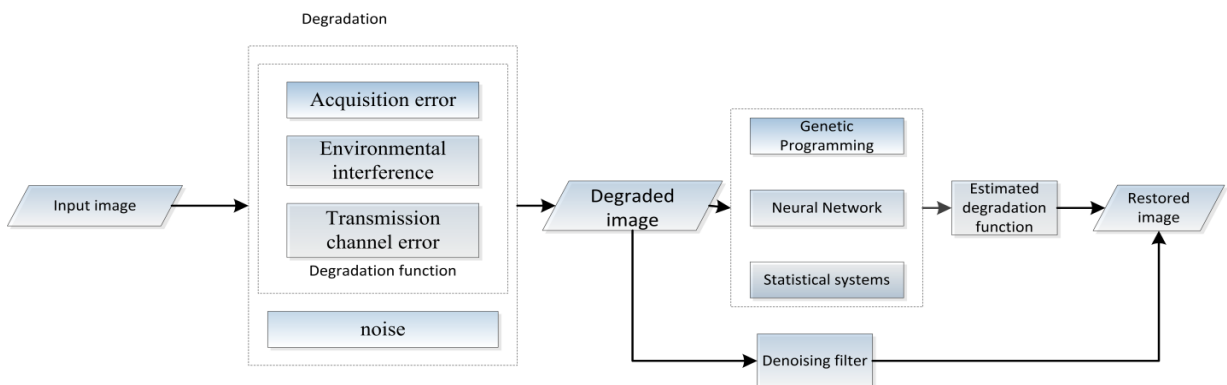

Figure 5: General block diagram of intelligent image restoration techniques

### 5.2.1. GP in Image Deconvolution

In literature, GP is rarely used for image deconvolution [21] [22]. Among the few reported methods, GP based blind image deconvolution filter was proposed by Majeed et al.[22]. In Majeed's technique, for a small neighborhood of each pixel of a degraded image, a set of feature vectors was formed. An estimator was then trained by exploiting GP based automatic feature selection ability, to select and combine useful features. The proposed technique was compared with Richardson–Lucy (LR) deconvolution and Wiener filtering approaches, and comparatively good results were reported in terms of *RMSE* and PSNR.

### 5.2.2. GP in Image Denoising

Many researchers used GP as an effective strategy to remove noise from an image [23–29]. Chaudhry et al.[24] proposed GP for restoring degraded images by evolving an optimal function that estimated pixel intensity. Their technique was a hybrid of GP and Fuzzy logic, which denoises gray level Gaussian noise images in the spatial domain. First, for deciding if a pixel needed to be rigged, mapping function based on fuzzy logic was used. Then, GP was applied to evolve an optimal pixel intensity-estimation function.

Another denoising method based on local-adaptive learning (for Gaussian and salt & pepper noise) method was reported by Yan et al. [25]. In the training stage of their process, clustering was used to classify the image based on similar local structures. Then GP was applied to determine optimal filters (which themselves were tree-like individuals) for each cluster. The function set was composed of Gaussian and bilateral filters, as well as arithmetic operators. An increased PSNR was reported in comparison to other local learning-based methods such as K-clustering with Singular-Value-Decomposition.

On the other hand, to remove Racian noise from Magnetic-Resonance-Imaging (MRI), an optimal composite morphological filter was generated via GP [26]. In their method, a GP individual performs morphological operations on the corrupted image to obtain an observed copy. RMSE of the feature sets for the degraded image and the observed image was used to calculate the fitness of each individual. For evaluation, a noisy image was filtered by the developed filter to obtain an estimated copy. Moreover, their method (in terms of RMSE and PSNR) was also compared with other techniques.

Another work for removing mixed/Gaussian noise using GP was reported by Petrovic et al. [27]. GP based two-step filter (each having its estimator), was used to remove the noisy pixels missed by the first detector.

Harding et al. [28]used Cartesian GP to evolve image filters and evaluated their fitness functions on a Graphics Processing Unit (GPU). The average error on each pixel was used as the fitness score. Majid et al. [29] employed GP to estimate optimal values of noisy pixels for impulse noise removal. Noisy pixels were detected first using the directional derivative; then, their costs were estimated using GP estimator by incorporating noise-free pixels. Feature vectors were constructed using noisy pixels with at least three neighboring noise-free pixels. Recently, Beltran et al. [30] used a GP based restoration technique to remove haze from images. During training, GP based estimators were evolved based on MAE. Beltran's method showed significant performance improvement when compared with the latest techniques. However, the quality of restored images was only evaluated against MAE and PSNR metrics.

## 5.3. GP in Image Registration

Image registration involves matching different images of the same scene, which are captured at various intervals, from different directions or by different sensors. One objective of image registration is to bring into line the images in such a way so that high-level processing can be executed.

Only a few researchers have employed GP for image registration. Chicotay et al. [31] presented GP based approach for massive size image registration, in which transformation *T* on an image mapped every pixel of the input image to a different pixel in the coordinate system of the referenced image. Mutual Information (MI) was used as a measure to search for a function that generated the highest value when there existed a maximum overlap between the referenced and the transformed image. RMSE was used to evaluate each individual. A comparison was made with the Scale-Invariant Feature Transform (SIFT) [32] based image registration. Though the results were not as good as a SIFT-based technique, they were still comparable, keeping in view that unlike the SIFT-based method, their technique did not make any assumptions about the transformation model to initiate or bound the registration process. The function set included transformation functions such as sine, cosine, power, rotation, and radial basis function.

Langdon et al. [33] employed GP optimization to improve GPU based implementation of Nifty Reg Software. Whereby the Nifty Reg is open-source software for medical image registration, and the optimization was performed for six different graphics cards. The implementation was carried-out using Compute Unified Device Architecture (CUDA). GP with linear variable-length genome specified changes to the CUDA kernel. Two parameters (compute level and size of the block) for CUDA were also tuned along with post-evolution bloat removal. Each genome was saved as a text line. Crossover and mutation were prohibited from including code lines to those parts of the kernel, where the containing variables might go beyond the scope. For each generation, a new image was created randomly, and each GPU kernel was run on it. GPU kernel generated an answer which was checked against that of the CPU. And the runtime was compared with that of the original Kernel.

Outliers within data significantly degrade the performance of a classifier. To overcome such degradation in the performance of an image registration classifier, Lee et al. [34] reported a novel GP based method. In their approach, firstly, feature extraction was performed using SIFT [32]. The features were then classified into three categories, i.e., inliers, outliers, and non-classified features. Inliers and outliers extracted from the first phase were provided as training data to GP. GP then categorized the non-classified features into two groups, i.e., inliers and outliers. All the outliers were removed from the dataset. And the image registration was performed on the pre-processed data (after outlier removal). The Block diagram of Lee's technique is shown in Figure 6.

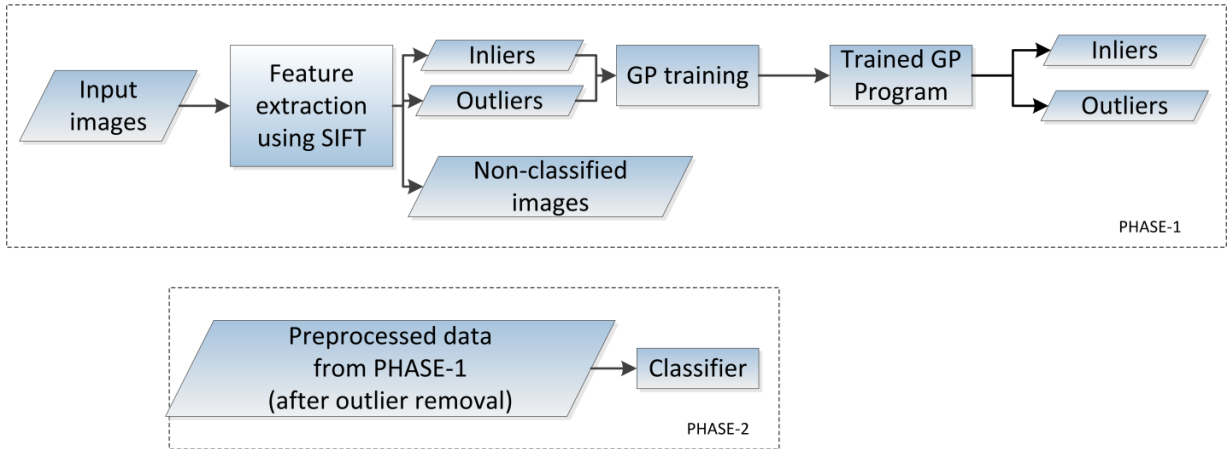

Figure 6: GP based Image Registration technique by Lee et al. [34]

### 5.4. GP in Image Compression

The increasing use of images and their storage requirements initiated the need to compress them. The basic idea behind image compression is to remove redundant bits, and thus encode the information contained in the image so that while restoring, the encoded image is obtained without considerable loss. Restoring the exact image is vital in case of medical diagnosis or other security forensics. Transmitting images over the internet also requires compression to consume less bandwidth.

Fukunage et al. [35] described a GP system for lossless image compression, which learned a nonlinear model for pixel prediction based on a pixels. Four adjacent pixels were used as terminals for the GP. For each image, a unique model was generated and was represented as *s-expression*. The high computational cost of evaluating the *s-expression* for each pixel was overcome by removing function call overhead by employing the Genome Compiler. This compiler translates *s-expressions* into efficient SPARC machine code before execution. The proposed method was compared with other compression techniques, including CALIC, LOCO-I, gzip, and was reported to be superior in the compression achieved, though it was slow. Figure 7 depicts the steps of Fukunage's method.

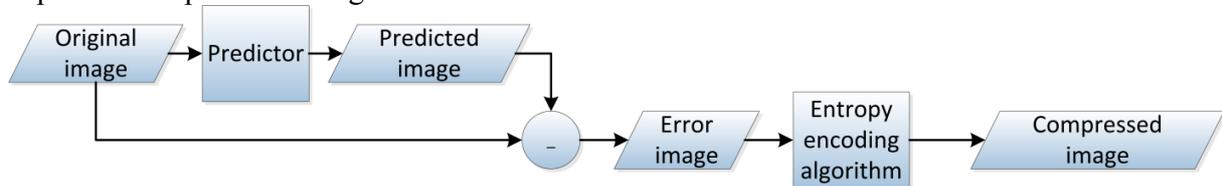

Figure 7: GP based Image Compression technique by Funkunage et al. [35]

In another technique, Parent et al. [17] proposed lossless compression of medical images. They used a linear GP driven by cGA and found a transformation, represented as *T(d)*, which improved the compression ratio of data *d*. Moreover, this transformation could remove certain types of redundancy. The terminal set comprised of constants, and the function set included four transformation functions. These transformations acted as preprocessing before real compression and yielded enhanced compression as compared to standard GA based techniques.

## 5.5. GP in Image Segmentation

The primary purpose of image segmentation is to segment out different gray levels of an image. If the pixels belonging to regions are homogeneous, then they are assigned the same label. Otherwise, various labels are attached. In other words, a good segmentation criterion is required to look for homogeneity within-region and heterogeneity between regions [36].

Developing a comprehensive way to check the accuracy of image segmentation algorithms is a significant problem. In the field of image processing, GP has been widely used to segment region of interest from images [37–46]. Vojodi et al. [40] used GP to combine different and unrelated evaluation measures. They selected three evaluation measures, which are based on the layout of entropy, similarity within the region, and disparity between the areas for the creation of composite evaluation measures.

In another technique, Song et al. [41] used GP to evolve automatic texture classifiers, which were then used for texture segmentation. As opposed to conventional methods, their method does not require the manual construction of models to extract texture features because the classifier's input is raw pixels instead of features. Also, the conventional methods are not universally applicable as they rely on the knowledge of the nature of texture, which may differ from region to region and image to image.

GP can capture variation within images; that is why GP is popular in evolving a suitable image segmentation technique. However, GP based techniques mainly developed vast and expansive segmentation algorithms. In this regard, Liang et al.[44] proposed a multi-objective GP based segmentation technique, in which classification accuracy and program complexity are included within the fitness function. Liang's method evolved a suitable solution with an optimal tradeoff between accuracy and program complexity. In another approach, the GP based segmentation technique developed an accurate and reliable figure-ground segmentation [45]. Their segmentation approach was evaluated against four different data sets.

Similarly, another segmentation technique was reported that used strongly type GP and used two-stage during the GP evolution cycle [46]. Dong et al. [42] attempted to categorize the texture within an image to be either Corpora Lutea (CL) (i.e., an endocrine gland that is generated from the follicular tissue after ovulation) or non-CL, based on local neighborhoods. A 16-bit invariant uniform Local Binary Patterns (LBP) histogram of pixels in the neighborhood was formed to represent texture descriptions. Feature vector created by the histogram bin values was fed as input to GP. GP was used to train a classifier for distinguishing between CL texture and other textures. For segmentation, a sliding window was used to scan the image in raster order. The GP classifier then assigned each image pixel in the window, a class label. Majority voting was used in the case of multiple tags. For CL detection, properties related to the set of the region were computed for each image's output region. Then a GP classifier was trained using these properties. Finally, the classifier was used to detect whether the segmented part of an image is a CL or not.

To address the tradeoff between localization accuracy (requiring a small window) and noise rejection (requiring large window) posed by selecting the window size, Fu et al. [47]used GP to automatically search discriminating pixels and their neighbors for constructing edge detectors. Rather than using a set of pixels from a moving window, GP used a full image. The selected pixels were then used to form linear and nonlinear filters for detecting edges. The parameters of these filters were estimated via a hybrid of Particle-Swarm-Optimization (PSO) and Differential

Evolution. A shifting function, representing four directional shifting functions, was included in the function set. A comparison was made with other detectors showing good results for GP based detectors. They employed *F*-measure to evaluate the accuracy of the detectors. Similarly, another GP based image segmentation technique for extracting regions of interest from the background was proposed by Liang et al.[43]. Feature selection using GP was used to find out the useful features that helped to segment out the desired region of interest. Three different types of GP based feature selection methods were proposed. In all of the three ways, fitness function within GP was either based on a single or multi-objective method. Their experimental results showed that the GP based feature selection, which used multi-objective fitness function, improved the performance of the classifier, and also reduced the computational complexity. The block diagram of Liang's technique is shown in Figure 8.

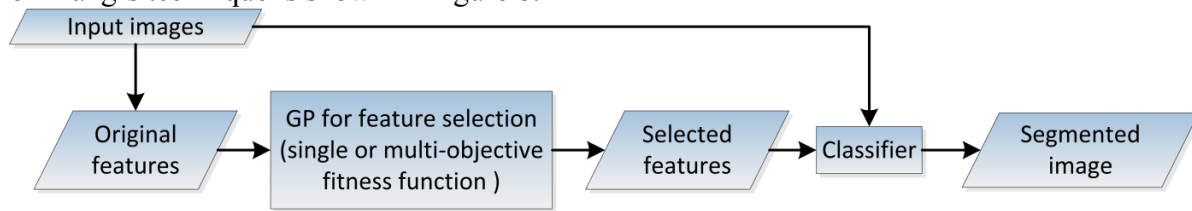

Figure 8: GP based image segmentation by Liang et al.[43]

## 5.6. GP in Image Retrieval

Due to the decline in the prices of image acquisition devices and the development of efficient image processing algorithms, the image databases are increasing in number. Consequently, it has become inevitable to design effective and fast methods for retrieving desired images from such significant collections. There are different techniques for image retrieval, such as associating some metadata (tags, keywords) with the images, or using content-based retrieval, which is based on similarities of the contents of the given image (or feature) and the desired image. Different shapes, textures, colors, etc., can be used as features for Image retrieval related tasks.

In an exciting work by Torres et al. [48], GP was applied for creating a merged similarity function for content-based image retrieval. They showed that features could be combined from multiple feature vectors, or weights can be assigned based on image similarities. In the case where combining images gets more complicated, then the GP is used for combining nonlinear image similarities. The resulting composite descriptor was simply a combination of pre-defined descriptors. This GP based composite descriptor combined the similarity values obtained from each descriptor and then produced a more effective similarity function.

Ciesielski et al. [49] used a segmentation algorithm based on texture-versus-all-else classifiers. These classifiers were evolved by GP to retrieve from an extensive heterogeneous collection of images.

Calumby et al. [50] used GP to iteratively combine multimodal similarity measures, such as those extracted from text and content, to generate new similarity functions that would fit the user preferences. For each discovered function, the evaluation returned a measure of quality that was based on how well that function ranked the training set objects. The proposed method showed higher efficiency when compared to Image CLEF Photographic Retrieval Task [51]. A somewhat similar framework was also reported by Ferreira et al. [52].

Saraiva et al. [53], on the other hand, used GP to combine multiple textual sources of evidence, such as image file name, the content of HTML, page title, alt tag, keywords, description, and text passages around the image, to rank web-based image retrievals.

## 5.7. GP in Image Classification

Image classification is the process of classifying images based on the visual contents. Various Artificial Intelligence (AI) based technologies, such as Artificial Neural Networks (ANNs) and fuzzy systems, have been applied to develop autonomous classification algorithms and have shown promising results [54].

Two broad families of machine learning approaches used in image classification are parametric (that requires learning phase) and nonparametric methods (that does not require learning phase). Some examples of parametric classifiers are Support Vector Machine (SVM), Decision Trees, and GA. Whereas, Nearest-Neighbor image classifier is an example of nonparametric classifiers. When GP is used for classification, the inputs are features, and the output is a mathematical expression that returns different values for different classes.

Using GP for classification requires a threshold to be set for the program output to specify different classes. In the case of static range selection, boundaries of program output space are fixed and predefined. However, in dynamic range selection, the boundaries are searched automatically [55]. In centered dynamic range selection, the class boundaries are dynamically determined by calculating the center of the program output values for each class. In the slotted dynamic class boundary determination method, the output value of a program is split into many slots. Each slot is assigned to a value for each class. It then dynamically determines the class by only taking the class with the highest value at the slot [56]. Several techniques have used GP for classification [57–63]. Nandi et al.[64] used GP for feature selection to classify breast masses in mammograms into benign and malignant groups. To narrow down the pool of features, they used a few procedures like Sequential Forward Selection and Student's t-test, etc. Once important features were selected, these were divided into two groups. Either union or intersection operation was performed over these groups to create a new set of data points for the GP classifier.

Similarly, Kobashigawa et al. [54] showed that with the increase in problem difficulty level, GP achieves better results than ANN methods. Kobashigawa's work also revealed the robustness of GP to unseen examples along with an inherent capability of optimal global searching, which could minimize efforts that are required during training processes. On the other hand, Smart et al. [56] employed the evolutionary process of GP to dynamically determine the boundaries between images of coins having different denominations. Pixel level domain-independent statistical features such as average intensity, variance, etc. were given as input to GP to automatically select features that were relevant to this multi-class image classification problem. As compared to a static range selection, reasonably good results were reported using the dynamic methods, centered dynamic range selection, and slotted dynamic range selection.

Similarly, Atkins et al.[65] proposed a GP-based domain-independent technique for extracting features and image classification. The Block diagram of Atkins's approach is shown in Figure 9. First raw images were preprocessed by the filtering layer whose outputs (the filtered images) were fed to the second layer, called the aggregation layer. The aggregation layer then performed feature aggregation and produced a real value. Finally, the output of the aggregation layer was passed on to the classification layer to perform classification. For this layer, a threshold of zero was used, so a negative output would mean class A and non-negative would classify the image as belonging to class B. The proposed procedure was tested on four different datasets, and the reported results suggested that it outperformed the basic GP methodology with increasing problem difficulty.

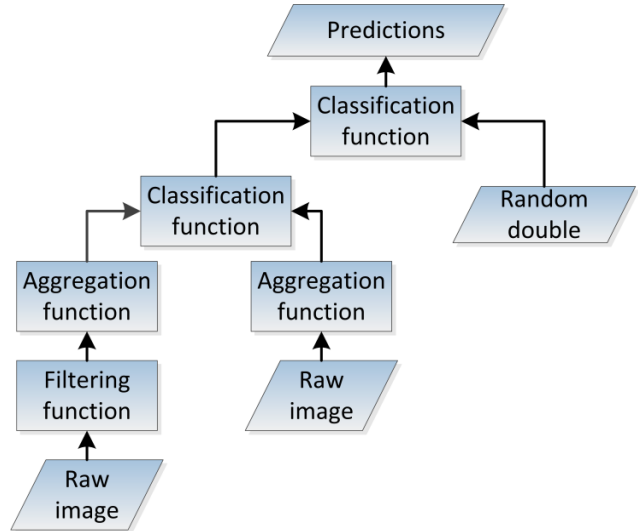

Figure 9: Three tier GP for image classification [65]

In another approach, Al-Sahaf et al.[66] presented a GP based approach that extended the work of Atkin's et al. [65] and introduced aggregation functions that read in different shapes such as lines, circles, and rectangles to enable sampling windows that were not square. They did not use the filtering layer and still achieved better results as compared to a canonical GP that used extracted features and performed classification by the three-tier GP. Guo et al.[67] used a Modified Fisher criterion-based GP (MF-GP) for generating features. The generated features were evaluated for their discriminating ability by the Minimum Distance Classifier (MDC). Improved results were reported for MF-GP compared to Multi-layer perceptron, SVM, and Alternative Fisher criterion-based GP (AF-GP) with MDC.

A semi-automatic approach for classifying Remote Sensing Images (RSI) was proposed by Santos et al. [68]. GP was used to learn user preferences via user indicated relevant as well as non-relevant regions. The image region descriptors were combined that encoded color and texture properties. The reported results showed that the method outperformed maximum-likelihood-classification when used for RSI classification. In the same way, Santos et al. [69] improved the results of the previous work by combining Optimum-Path-Forest (OPF) with composite descriptors obtained using the GP framework. OPF classifier represents each class of objects by one or numerous optimal-path trees rooted at essential samples, called prototypes. The OPF-based classification system took into account user interaction.

Choi et al. [70] proposed a system for automatic detection of pulmonary nodules, which first segmented the lung volume using thresholding, then detected and segmented nodule candidates using multiple thresholding and rule-based pruning. From these nodules, geometrical and statistical features were extracted, and a GP-based classifier was trained. The fitness function was constructed by combining the AUC-ROC curve, the True-Positive-Rate (TPR), and specificity. They reported that compared to the previously proposed methods for this application, this GP based classifier showed high sensitivity and a reduced false-positive rate. Zang et al.[71] developed fitness function for classification based on probabilities (derived from Gaussian distribution) that are associated with different classes. Two fitness functions (using the overlapped region and weighted distribution distance) were developed, assuming the outputs from different classifiers as random independent variables. Zang's approach exploited many top GP programs for classification, and the class with the highest probability was used as the class of the object pattern. In comparison to a primary GP classification, which also used multiple best programs and voting, the proposed technique was reported to have good results in terms of classification accuracy and execution time. Recently, Cava et al. [72] proposed GP based multi-class classification technique in which transformation is performed against multi-dimensional feature space. Cava's technique was compared to different techniques related to different domains. Results showed that the proposed solution could scale well to those problems that have high feature dimensions. Similarly, Burks et al. [73] proposed a GP based classification technique for detecting tuberculosis from X-ray images. Unlike traditional classification methods, proposed work no pre-processing and segmentation steps before the training of the GP based classification approach. Moreover, Burks's technique needs less training time in comparison to other traditional proposed methods.

### 5.8. GP in Image Watermarking

The consistently broader use of information technology demands the protection of information. Therefore, in this regard, digital watermarking is used as a promising technique to overcome the issues related to the protection of information, especially for the authentication of medically related information. One of the measures to evaluate the quality of the watermarked image is to evaluate its imperceptibility. In watermarking techniques, imperceptibility shows that the visual appearance of the watermarked image should be close to the original image. However, when more information (payload) is embedded in the image, it causes distortion in the original image. That is why there is a tradeoff between imperceptibility and payload. In the past, many GP based watermarking techniques [74–81] have been proposed for the development of efficient and reliable watermarking systems. To increase the robustness and imperceptibility in digital image watermarking, GP was employed by Golshan et al. [75]. Instead of setting the Perceptual Shaping Function (PSF) to a constant function, GP was utilized to develop an intelligent PSF. A fitness function based on both robustness and imperceptibility was used to evaluate the performance of each PSF individual. Similarly, Golshan et al. [76] used a hybrid approach of GP and PSO for the same purpose. In technique, developed by Gilani et al. [77], GP was used to estimate the distortion within the distorted watermarked signals. Both the watermarked and the distorted watermarked signals were fed to a GP module. The best-estimated distortion function returned by GP was then applied to the original watermarked signal. Varying strengths of Gaussian and JPEG compression attacks were tested for the proposed technique.

Similarly, Usman et al. [78] proposed evolving application specific Visual Tuning Function (VTF), in which GP optimizes the balance between imperceptibility and robustness while processing an 8x8 block of Discrete Cosine Transform (DCT) image. The watermark was structured according to the Human Visual System (HVS) and a cascade of attacks. VTF is given as:

$$\alpha_G(k_1, k_2) = f(X_{0,0}, X(i, j), \alpha(i, j)), \tag{1}$$

where $X_{0,0}$ is the discrete cosine coefficient and signifies dependency of VTF on luminance sensitivity, $X(i, j)$, is AC coefficient and symbolizes dependency of VTF on contrast masking, and $\alpha(i, j)$ shows frequency sensitivity. The current value of Watson's VTF, DC, and AC (DCT) coefficients of 8x8 blocks were provided as variable terminals. Each potential VTF was evaluated for imperceptibility related fitness, whereas for robustness, Bit Correct Ratio (BCR) represented an objective measure. Test images were then watermarked with the evolved VTF.

To select the watermarking level, Jan et al. [79] proposed GP based approach. Coefficients were selected using a 32x32 block, whose Discrete Wavelet Transform (DWT) was obtained. Luminance, contrast, and Noise-Visibility-Function (NVF) were used as terminals for GP trees. Watermarking level was given by:

$$\alpha = f(\text{lum}(i, j), \text{cont}(i, j), \text{co}(i, j), \text{NVF}(i, j)), \qquad (2)$$

where *co* is a selected coefficient, *cont* is contrast, and *lum* is luminance. Robustness against different attacks was reported, whereas to check the imperceptibility of the watermark, Mean Square Error (MSE) and PSNR were used. Similarly, Abbasi et al. [80] used a similar approach but used a block size of 4x4. Khan et al. [81] presented a DCT based watermarking system that employed GP for finding optimal perceptual shaping function according to HVS. Each GP tree represented a perceptual shaping function, which was evolved to embed high strength watermark in areas of high variance and low strength watermark in areas of low variance. Change in a local variance of the watermarked image with respect to the original image was used as a fitness function. This technique was tested for JPEG compression and Gaussian noise. Recently another interesting reversible watermarking technique based on GP for the protection of medical-related information was proposed by Arsalan et al. [74]. The Block diagram of Arsalan's technique is shown in Figure 10. First, the histogram modified image was formed after the pre-processing of the original image. Integer Wavelet Transform (IWT) was then applied to the histogram modified image. After applying IWT, GP was used to find out the coefficients within the wavelet domain for the purpose of embedding watermark. The aim of the proposed GP based intelligent watermarking scheme was to produce a watermarked image having low distortion and high payload.

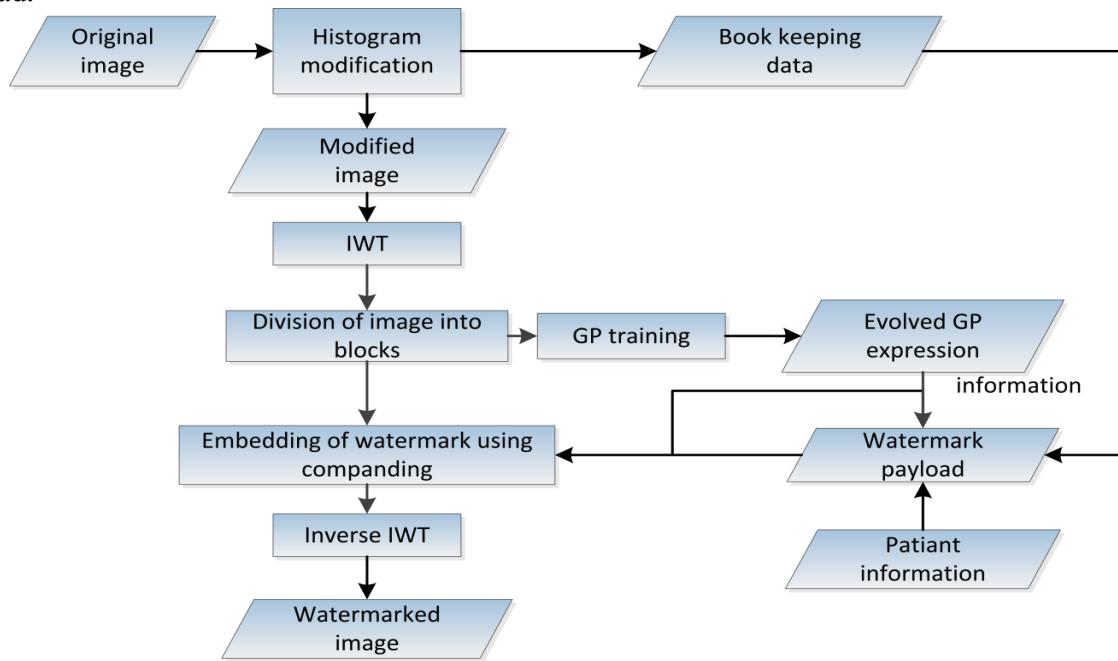

Figure 10: GP based watermarking technique by Arsalan et al. [74]

## 5.9. GP in Object Detection

Object detection is the task of finding different types of objects belonging to different categories and is a challenging task especially, in the field of image processing and computer vision. In the field of image processing, GP has been used by many researchers [82–95] for accurate and efficient prediction of objects from cluttered and noisy scenes or images. In a review article, Krawiec et al. [96] analyzed the applications of GP in object detection related applications.

Howard et al. [83] utilized GP to evolve detectors to detect ships in Synthetic Aperture Radar (SAR) imagery. Terminal nodes were real numerical values derived from random constants or pixel statistics. A value greater than zero was decided to be a target detection, while a value of zero or less was for ocean pixel. In Lin et al. approach [88], GP was used to synthesize composite operators and features from primitive operations and features for object detection. A composite operator was applied to primitive feature images; the output was segmented to obtain a binary image and was used to extract the target object from the original image. The size of a composite operator, as well as misclassified pixels, were taken into consideration, while the fitness function used in Lin's technique was based on the Minimum Description Length (MDL) principle. In another work, Bhanu's et al. [89] have used a similar approach of composite operators, but instead of MDL based fitness function, they used the following fitness measure:

$$Fitness = \frac{n(G \cap G')}{n(G \cup G')}, \qquad (3)$$

Where, $G$ and $G'$ are foregrounds in the ground-truth and in the detected image respectively, $n$ being the number of pixels in a given region.

Martin et al.[90] used GP to create algorithms for obstacle detection, which analyzes a domain to find its constraints. Lowest non-ground pixels were manually marked, and these images were fed to GP, whose output was then compared to the ground truth images. A robot was then controlled by the best-evolved program.

Edges are detected traditionally by using local window filters; however, in Fu et al.[91] work, GP was used for domain-independent global edge detection using the whole raw image as input. Different shifting functions were used along with other commonly used operators. F-measure was used in constructing the fitness function:

$$F_m = \frac{1}{N} \sum_{i=1}^{N} (1 - \frac{2 r_i p_i}{r_i + p_i}), \qquad (4)$$

where $i$ represents image, $N$ the number of images, $r_i$ and $p_i$ are the recall and precision of a given image. Better results were reported as compared to Laplacian and Sobel edge detectors.

In another work, Fu et al. [92] used GP to evolve edge detectors. Instead of distributing a fixed size window into small areas based on different directions, it searched for features based on flexible blocks, and the fitness function was based on F-measure. Similarly, GP was also used for improving the performance of an edge detection system, where the fitness function was based on the accuracy of the training data [93]. In another work by Fu et al. [94], composite features were constructed for edge detection by estimating the observations of the programs evolved by GP as triangular distributions. Gaussian filter gradient, histogram gradient, and normalized standard deviation were used as a terminal set. In order to detect edges, an unsupervised GP system was proposed in [95]. However, fitness function was based on the energy functions in the active contours. In comparison with a Sobel edge detector, the evolved GP edge detectors were reported to have better performance.

Similarly, Liddle et al. [84] used a Multi-Objective GP (MOGP) for object detection. MOGP evolves a set of classifiers rather than a single classifier classifier, as in the case of Single-Objective GP (SOGP). The proposed technique used the NSGA-II algorithm, whose performance measure is Non-Dominance-Ranking and Crowding Distance. A two-phase training process applied the MOGP algorithm twice using different objectives e.g., maximizing both TPR and True Negative Rate (TNR); or maximizing Detection Rate (DR) while at the same time minimizing False Alarm Rate (FAR). In the interesting work of Zang et al. [89], GP was used for object detection, but instead of using raw pixels and terminals, they used pixel statistics such as mean, standard deviation, and moments. A new fitness measure termed as "false alarm area" was used along with a combination of DR and FAR.

On the other hand, Zang et al. [82] presented domain-independent features such as mean and standard deviation as terminals for GP to detect multiple objects. They used three different ways (rectilinear: based on different rectangles; circular: using circles of different radii, and using an average of pixels) for obtaining pixel statistics. Evaluation of programs was performed with a fitness function based on DR and FAR as such:

$$fitness = K1 \cdot (1 - DR) + K2 \cdot FAR, \qquad (5)$$

where, K1 and K2 are constants. Zang et al. [86] introduced a two-phase GP approach for object detection. In the first phase, cutouts from the training images were used with classification accuracy as the fitness function. The second phase was initialized with the population from the first phase, and a window was moved over the whole image. For the second phase, the following fitness function was used:

$$fitness = K1\,(1-DR) + K2 \times FAR + K3 \times FAA + K4 \times size \qquad (6)$$

where *FAR* is the false alarm rate, *DR* stand for detection rate, *FAA* is the false alarm area (positive classifications − objects in the image), *size* is the program size, while *K1, K2, K3, and K4* are constants.

Hunt et al. [87] followed the previous two-phase approach [86], augmented with validation and sampling methods in order to avoid overfitting. Validation was performed after every two generations. Generalization ability is usually evaluated by calculating the hyper-area (area covered by the best Pareto-front), and distance (the difference between the performance of the classifier on training and validation set). Nguyen et al.[97] used GP for the detection of rice leaf. In Nguyen's work, the dataset was created by taking images from the top of the rice field, and a total of 600 images of size $640 \times 840$ were captured from the camera. Out of the total 600, 300 images were used for the training of the classifier. After capturing images, the next step was the conversion of color images into grayscale. The below equation shows the conversion of colored images into the grayscale images.

$$Pixel(gray) = 0.3 \times \mathrm{Re}d\_channel + 0.59 \times Green\_channel + 0.11 \times Blue\_channel \qquad (7)$$

In order to deduce the positive and negative samples from the set of gray images, a window size of $20 \times 20$ pixels was used to extract sub-regions within the images. If each sub-image contained a portion of rice leaf then, it was labeled as a positive example; otherwise, the negative label was assigned to that subpart. After pre-processing of original images, a total of 9000 images of size $20 \times 20$ pixels was generated in which half belonged to a positive class, and half belonged to the negative class. For training of GP program, pixels were considered as a terminal set, whereas the function set was comprised of four different arithmetic operators and a square-root function. The weighted sum of TPR and TNR was used as a fitness criterion. In order to ensure that the value of fitness was between 0 and 100 percent, the following constraint was followed $w1+w2=1$. The block diagram of Nguyen's technique is shown in Figure 11.

$$Fitness = w1 \times TPR + w2 \times TNR \qquad (8)$$

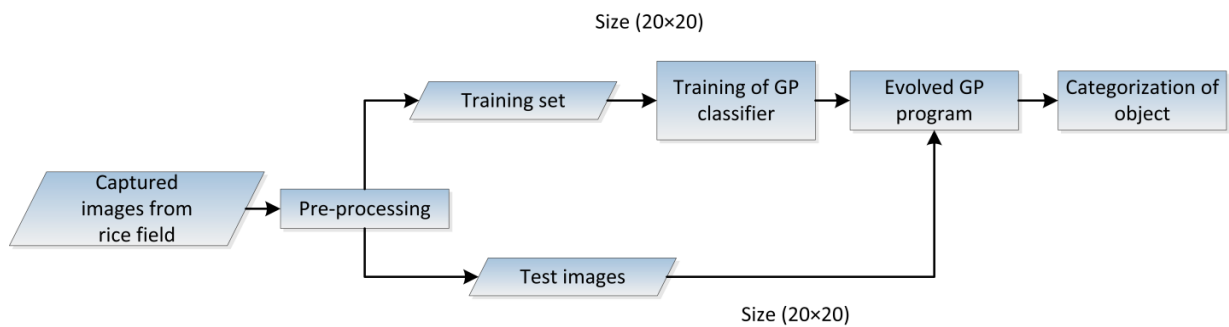

Figure 11: GP based object detection technique by Nguyen et al.[97]

## 5.10. GP in motion detection

In the past, many modeling and background subtraction related techniques have been designed for motion detection. Moreover, to avoid manually coded motion detection systems, different researchers used GP based automatically evolved systems [98–103]. It was observed that generally, the GP based evolved programs outperformed manually coded programs. To tackle the unstable background (such as rainy background, moving background due to a moving camera) in motion detection, GP was employed in [104], where classification accuracy based on motion and non-motion was used as a fitness measure.
Another difficult task in the case of motion detection is to detect motion from a noisy scene when there is no information about the noise. Pinto et al. [100] tackled this problem by using GP based approach in which motion detectors were generated during the testing phase on the basis of the fitness function. In this approach, Gaussian noise was added in the video [100] and showed better results for detecting motion in different environments. In another work [105], the GP program was used for analyzing the various type of motion detection techniques such as detecting simple motion, detection of fast-moving objects, motion detection from a noisy background. Another advantage of using GP for motion detection is that the evolved detectors can also tolerate noise; that is why GP may be considered as one of the best approaches for the detection of motion.
Similarly, Xie e al. [98] used GP for anomaly detection from crowded scenes. In Xie's approach, multi-frame Local Binary Patterns (LBP) difference based on LBP was used for extracting features from video frames. Training of GP was performed on extracted features. The proposed scheme detected abnormalities in real-time videos. Similarly, Song et al. [106] proposed GP based target motion detection approach that automatically evolved GP program and separated target motion from other irrelevant motions such as the noisy background. The technique proposed by Song et al.'s was comprised of two phases. In the first phase (evolution phase), the data used during training was divided into training and test parts. Parameter optimization during training was performed on the basis of the performance of GP based evolved programs on test data. After the evolution of the GP program, the next phase was the application phase, in which the best-evolved GP program from the evolution phase was used to check the performance on unseen data samples. The block diagram of Song's technique is shown in Figure 12. This technique was used for detecting motion from the video, so the first two-dimensional array of size 20 x 20 was captured as video frames from different locations of videos. The image was assigned to the positive class if the majority of pixels within the frame were labeled as sampled by a human expert. During the GP training, program accuracy was used as a fitness function, whereas detection accuracy versus the number of generations was used as an evaluation measure.

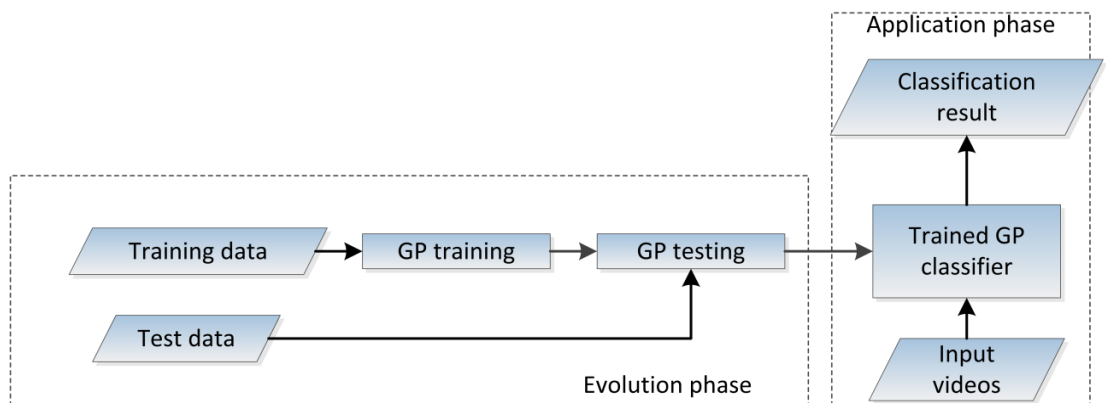

Figure 12: GP based motion detection technique by Song et al. [106]

## 6. Category wise Applications of GP

This section presents different GP based techniques that are applied to different categories of image processing. Table 1 lists the references as well as the GP parameter settings for each category. An overall analysis of Table 1 shows that in all of the reported image processing related applications, a large population is used in comparison to the number of generations. The large population within each generation helps to increase diversity and hence increases the chance to obtain better individual in less number of generations. Moreover, most of the GP related image processing applications used tournament selection. The advantage of using the tournament selection method is that it helps to maintain constant selection pressure, and even programs with average fitness have chances to reproduce a child in the coming generation. Table 1 shows that a higher crossover probability is used in comparison to mutation probability because higher values of mutation probability increase the search area within the search space, and the algorithm may get stuck in local minima. Also, in image processing related applications, ramped half and half is the commonly used population initialization method. This method produces the initial tree of variable length and thus help to increase the diversity of the initial population. The last column of Table 1 shows that several runs are carried out in most of the reported works to show the effectiveness of proposed methods.

| Category | References | Number of generations | Population size | Selection method | Mutation rate | Crossover rate | Population initialization methods | Runs |
|---|---|---|---|---|---|---|---|---|
| Enhancement | [19] | 100 | 4096 | ----- | 0.25 | 0.5 | Random growth | ----- |
| Restoration | [25] | 50 | 500 | Tournament | 0.05 | ----- | Ramped half-and-half | ----- |
|  | [26] | 200 | 200 | Lexicographic tournament | Variable | ----- | Ramped half-and-half | ----- |
|  | [27] | 300 | 100 | ----- | ----- | ----- | ----- | ----- |
|  | [28] | ----- | 50 | Tournament | 0.05 | Not used | ----- | ----- |
|  | [29] | 500 | 50 | Tournament | Variable | Variable | Ramped half-and-half |  |
| Registration | [31] | ----- | 150 | Elitism | 0.3 | 0.9 | ----- | ----- |
|  | [33] | 500 | 300 | 50% truncation | 0.5 | 0.5 | Random single mutants | ----- |
|  | [34] | 50 | 300 | Tournament | Variable | Variable | Ramped half-and-half | ----- |
| Compression | [35] | 30 | 500 | Tournament | Not used | 0.9 | ----- | ----- |
|  | [17] | 50 | 500 | Fitness proportional | 0.05 | 0.8 | Random | 100 |
| Segmentation | [47] | 200 | 500 | ----- | 0.15 | 0.8 | ----- | ----- |
|  | [40] | 25 | 100 | Tournament | 0.2 | ----- | Ramped half-and-half | ----- |
|  | [42] | 500 | 600 | Tournament | 0.1 | 0.8 | Random | 3-folds |
|  | [43] | ----- | 500 | ----- | 0.1 | 0.9 | ----- | ----- |
| Retrieval | [49] | 150 | 200 | Fitness proportional | 0.0 | 0.9 | ----- | ----- |
|  | [48] | 25 | 600 | Tournament | 0.25 | ----- | Ramped half-and-half | ----- |
|  | [50] | 20 | 60 |  | 0.2 | 0.8 | ----- | ----- |
|  | [52] | 10 | 60 | Tournament | 0.2 | 0.8 | Ramped half-and-half | 10 |
|  | [53] | 30 | 300 | Tournament | 0.05 | 0.9 | Ramped half-and-half | ----- |
| Classification | [64] | 500 |  | ----- | ----- | ----- | ----- | 100 |
|  | [54] | 30 | 500 | ----- | ----- | ----- | Random | ----- |
|  | [56] | 50 | 300 | Fitness proportional | 0.3 | 0.5 | Ramped half-and-half | 10 |
|  | [59] | 50 | ----- | Tournament | 0.2 | 0.8 | Ramped half-and-half | 30 |
|  | [60] | ----- | ----- | Roulette wheel | 0.3 | 0.8 | Ramped half-and-half | 10 |
|  | [61] |  | 100 | Tournament | 0.19 | 0.80 | Ramped half-and-half | 30 |
|  | [65] | 50 | 1024 | Tournament | 0.29 | 0.8 | ----- | 40 |
|  | [66] | 50 | 1024 | Tournament | 0.19 | 0.8 | Ramped half-and-half | 30 |
|  | [67] | ----- | ----- | ----- | ----- | ----- | Random | ----- |
|  | [68] | 10 | 60 | Tournament | 0.2 | 0.8 | Ramped half-and-half | ----- |
|  | [70] | 80 | 300 | Generational | Variable | Variable | Ramped half-and-half | ----- |
|  | [71] | 51 | 500 | Fitness proportional | 0.3 | 0.6 | Ramped half-and-half | 50 |
| Watermarking | [74] | 50 | 25 | Roulette | Variable | Variable | Ramped half-and-half | ----- |
|  | [75] | 100 | 10 | Keep best | 0.1 | 0.9 | ----- | ----- |
|  | [76] | 100 | 10 | ----- | 0.1 | 0.9 | ----- | ----- |
|  | [77] | 32 | 120 | Tournament | Variable | Variable | Ramped half-and-half | ----- |
|  | [78] | 40 | 160 | Tournament | Variable | Variable | Ramped half-and-half | ----- |
|  | [79] | 10 | 25 | ----- | ----- | ----- | ----- | ----- |
|  | [81] | 30 | 300 | ----- | ----- | ----- | Ramped half-and-half | ----- |
| Object detection | [83] | 40 | 1000 | Tournament | Not used | Unknown | Ramped half-and-half | ----- |
|  | [88] | 70 | 100 | Tournament | 0.05 | 0.6 | ----- | ----- |
|  | [90] | 51 | 4000 | Tournament | ----- | ----- | Ramped half-and-half | ----- |
|  | [91] | 250 | 200 | ----- | 0.15 | 0.8 | ----- | 30 |
|  | [92] | 200 | 600 | ----- | 0.15 | 0.8 | ----- | 30 |
|  | [93] | 200 | 500 | Elitism | 0.15 | 0.80 | ----- | 30 |
|  | [94] | 200 | 200 | ----- | 0.15 | 0.8 | ----- | 30 |
|  | [95] | 100 | 30 | ----- | 0.3 | 0.65 | ----- | 30 |
|  | [84] | 60 | 500 | Tournament | 0.3 | 0.7 | ----- | 40 |
|  | [89] | 70 | 100 | Tournament | 0.05 | 0.6 | ----- | 10 |
|  | [86] | ----- | ----- | ----- | ----- | ----- | ----- | ----- |
|  | [82] | 100, 150, 150 | 100, 500, 700 | Fitness proportional | 0.25 | 0.70 | Ramped half-and-half | 10 |
|  | [87] | 20, 40 | 500 | Tournament | 0.30 | 0.70 | Ramped half-and-half | 40 |
| Motion detection | [99] | 300 | 30 | Elitism | 0.05 | 0.85 | ----- | ----- |
|  | [100] | 300 | 30 |  | 0.05 | 0.85 | ----- | ----- |
|  | [101] | 70 | 200 | Lexicographic parsimony pressure | 0.1 | 0.9 | Ramped half-and-half | ----- |
|  | [102] | 200 | 200 | Elitism | 0.1 | 0.85 | ----- | ----- |
|  | [103] | 100 | 50 | Tournament | Adaptive | Adaptive | Ramped half-and-half | 25 |
|  | [104] | 300 | 30 | ----- | 0.05 | 0.85 | ----- | ----- |
|  | [105] | 30 | 200 | Generational | 0.05 | 0.85 | ----- | ----- |
|  | [106] | 300, 150 | 30 | Elitism | 0.1 | 0.85 | ----- | 10 |

Table 1. Analysis of GP applications in image processing

## 7. Advantages and disadvantages of using GP for image processing

GP is a relatively new technique among all the evolutionary computing algorithms and has been widely applied in various image processing related techniques. In literature, GP has shown excellent performance for optimization and classification related problems; however advantages and disadvantages are also associated with GP based optimization techniques. Some of which are discussed below.

### 7.1. Advantages:

**Understandability:** GP outputs a program or a collection of programs in the form of mathematical expressions, which are easy to comprehend if simplified and converted to normal notation.

**GP vs GA:** Being the prominent types of Evolutionary Algorithms (EAs), both the paradigms share some characteristics but differ in others. They mainly differ in the way individuals are represented. GP uses a tree representation, whereas GA uses a string representation [107]. In the case of GA, individuals are generally raw data, whereas, in GP, the individuals are computer programs. The tree-based representation gives GP an edge over GA because of its flexibility; however, GA is faster compared to GP [108].

**Diverse Search Space:**

Genetic operators (crossover and mutation) used in GP introduce diversity and thus increases the span of search space. Larger search space helps in finding the most optimal solution for the problem at hand.

**Small Testing or Execution Time**

The GP needs considerable time for training an optimal GP based classifiers, but the finally selected GP tree needs very short execution time during the test phase. As GP based classification requires the small duration of test time, therefore, such systems are suitable for those applications in which appropriate time is available during training, and a short time is required during testing.

**Flexibility of GP Fitness Function**

Another advantage of using GP is that its fitness function is flexible and can be adjusted or designed according to the problem at hand. Moreover, multi-objective fitness functions are mostly used in image processing related tasks.

### 7.2. Disadvantages

**Computational Cost:** Fitness of each individual/program in the population is evaluated after every generation; therefore, the training process usually takes a long time. This shortcoming is considerably mitigated by recent advancement in CPU speed and number of cores, especially by using Graphical Processing Unit (GPU) [109].

**Needs Large Training Data:** A large dataset is needed for the training process in order to reach an optimal solution.

**No Guaranteed Solution:** Due to the stochastic nature, GP does not guarantee an exact solution; therefore it cannot be applied in situations where an exact straightforward solution is required.

## 8. Conclusion

This work presented a detailed study of the various image processing applications of GP. The automatic problem-solving capability of GP and increasing demand for image processing in a variety of fields has prompted researchers to look for efficient, robust, and cost-effective intelligent techniques. Moreover, due to the different nature of the image processing tasks, no hard and fast rules can be set. In addition, the terminal and function sets need to be problem tailored, and different fitness measures have to be developed. Also, by incorporating the domain knowledge related to the image processing field, GP is able to handle complex image processing tasks. In this paper, the application of GP in image processing related applications, different features of GP such as terminal and function set, fitness function, and other related parameters are discussed. Additionally, the Pros and cons of applying GP in image processing are discussed. Below are our observations related to applications of GP in image processing:

- In most of the applications of GP in image processing, large population size and crossover probability are used in comparison to the number of generations and mutation probability, respectively.
- In image processing related applications, the terminal set of GP is mostly set according to some statistical features related to the image.
- Tournament and Ramped half-and-half methods are used as a selection and population initialization method in most of the reported works.
- Selection of fitness function for a particular image processing application is the most important part and should be set in consultation with the expert of that image processing application.
- As parameter setting is also an important step in applying GP in any of the image processing related tasks. Before setting the GP parameters, a researcher must study and analyze the GP parameter settings in related image processing applications. This can help save time, whenever parameters of GP are needed to be set for any image processing related application.
- In literature, most of the reported work related to GP is oriented towards classification and object detection tasks.
- Relatively less work has been reported for image enhancement, registration, and compression, so more interesting techniques related to these fields can be exploited.
- Due to the heavy processing involved in image processing tasks, the algorithms require large training time. Training time can be considerably reduced by harnessing GPUs for enhanced algorithms.
- A GP based ensemble is likely to better exploit the decision spaces of the individual classifiers.
- Recently, Deep Neural Networks have shown remarkable performance in many Image Processing applications [110]. In this regard, the Meta classification/regression of individual learners, and specifically that of deep neural networks through GP, has good potential in learning complex problems[111].